\documentclass{amia}
\usepackage{lipsum} %Remove if not needed
\usepackage{booktabs}
\usepackage{array}
\usepackage{caption}
\usepackage{longtable}
\usepackage{graphicx}
\graphicspath{{./figures/}}
\usepackage{amsmath}
\usepackage{amssymb}
\usepackage{booktabs}
\usepackage{array}
\usepackage{float}
\usepackage{booktabs}
\usepackage{tabularx}
\usepackage{caption}

\begin{document}

%\title{Title of American Medical Informatics Association Submission}
%\title{Predicting Post-Acute Sequelae of SARS-CoV-2 in Adult Women Using LLMs with Wearable Data}
\title{Predicting Trajectories of Long COVID in Adult Women: The Critical Role of Causal Disentanglement}
%\author{Firstname A. Lastname, MD, MPH$^1$, Firstname B. Lastname, MD, PhD$^2$ }
%
%\institutes{
%    $^1$ Institution, City, MA; $^2$Institution, City, CA
%}
\author{Jing Wang$^1$, Ph.D., MD, Jie Shen$^2$, Ph.D.,  Yiming Luo$^3$, M.D., Amar, Sra$^4$, M.D., Qiaomin Xie$^5$, Ph.D.,  Jeremy C. Weiss, MD, PhD$^1$ }

\institutes{
	$^1$ National Library of Medicine, Bethesda, MD; $^2$Stevens Institute of Technology, Hoboken, NJ;
	$^3$Columbia University, NYC, NY;$^4$The George Washington university, Washington, DC
	$^5$ University of Wisconsin–Madison, Madison, WI 
}

\maketitle

\section*{Abstract}

Early prediction of Post-Acute Sequelae of SARS-CoV-2 severity is a critical challenge for women's health, particularly given the diagnostic overlap between PASC and common hormonal transitions such as menopause. Identifying and accounting for these confounding factors is essential for accurate long-term trajectory prediction. We conducted a retrospective study of 1,155 women (mean age 61) from the NIH RECOVER dataset. By integrating static clinical profiles with four weeks of longitudinal wearable data (monitoring cardiac activity and sleep), we developed a causal network based on a Large Language Model to predict future PASC scores. Our framework achieved a precision of 86.7\% in clinical severity prediction. Our causal attribution analysis demonstrate the model's ability to differentiate between active pathology and baseline noise: direct indicators such as breathlessness and malaise reached maximum saliency (1.00), while confounding factors like menopause and diabetes were successfully suppressed with saliency scores below 0.27.
 \section*{Introduction}

The persistence of debilitating symptoms following acute SARS-CoV-2 infection, clinically termed Post-Acute Sequelae of SARS-CoV-2 has emerged as a global public health crisis \cite{national2024long}. PASC is a multisystemic disorder characterized by a wide array of symptoms, including profound fatigue, cognitive impairment (``brain fog''), and respiratory distress, which can persist for months or years after the initial viral clearance \cite{thaweethai2023development}. Recent epidemiological data suggests that the prevalence of PASC is significantly influenced by demographic and biological determinants \cite{fritsche2023characterizing}. Specifically, women between the ages of 40 and 54 exhibit the highest risk profiles, with hormonal transitions such as menopause appearing to further exacerbate this susceptibility. Research indicates that women experiencing menopause within this age bracket face an approximately 42\% higher risk of developing PASC compared to age-matched men \cite{shah2025sex}, highlighting a critical need for sex-specific and endocrine-aware diagnostic frameworks.

A major challenge in PASC research is the ``signal-to-noise'' problem: many hallmark symptoms of Long COVID, such as insomnia, joint pain, and vasomotor instability, overlap significantly with symptoms of common comorbidities or natural biological transitions like menopause. These ``confounders'' can lead to diagnostic ambiguity, where environmental or baseline health factors are misattributed to viral pathology \cite{amin2021causation}. Previous causal inference studies have emphasized that failing to account for such confounding variables can severely bias clinical outcome predictions and treatment pathways \cite{hernan2010causal,yinrecipe}. For instance, sleep architecture changes and resting heart rate fluctuations are sensitive to both PASC and general aging, requiring a model that can disentangle true causal disease markers from baseline physiological noise.

The advent of wearable technology provides a unique opportunity to capture these physiological shifts in real-time \cite{straus2023utility,baigutanova2025continuous}. By monitoring cardiac activity (e.g., heart rate variability), respiratory metrics (SpO$_2$), and sleep architecture, we can move beyond self-reported surveys to objective, longitudinal data. However, the high dimensionality and inherent ``noise'' of wearable data require advanced computational approaches \cite{shen2022metric,shen2021sample,wang2018provable}. Large Language Models (LLMs) have recently shown promise in clinical settings due to their ability to process complex, multi-modal narratives and extract latent patterns from unstructured or semi-structured data \cite{moor2023foundation}.

In this study, we develop a specialized LLM-based framework to predict PASC severity three months in the future. Utilizing a cohort of 1,155 women from the NIH RECOVER database, our model integrates historical PASC score, clinical comorbidities, such as cardiovascular and mental health conditions, and four weeks of mean-aggregated wearable data. We specifically implement a causal-disentangled architecture to differentiate between ``Active Pathology'' symptoms and noises such as menopause-related symptoms, aiming to provide a more stable and clinically salient prediction of patient well-being. By normalizing and encoding these variables into a standardized semantic schema, we ensure that the model identifies true invariant features of PASC progression, even in the presence of significant clinical confounding.

\section*{Result}
To ensure the robustness and reproducibility of our findings, we employed a multi-seed validation strategy. The cohort was partitioned into training and testing sets using an 80/20 stratified split. To account for potential variability in model initialization and data distribution, the entire experimental pipeline, including data partitioning, model training, and evaluation, was repeated five times across five distinct random seeds ($S \in \{42, 100, 2024, 555, 777\}$). All reported performance metrics represent the mean and standard deviation ($\mu \pm \sigma$) across these five independent runs.

\begin{table}[h]
	\centering
	\caption{Comparison of Model Performance Metrics.}
	\label{tab:model_comparison}
	\begin{tabular}{lcc}
		\hline
		\textbf{Metric} & \textbf{XGBoost} & \textbf{Causal Inference Network} \\ \hline
		MSE             & $22.70 \pm 1.51$ & $30.01 \pm 13.36$                \\
		RMSE            & $4.76 \pm 0.16$  & $5.35 \pm 1.17$                  \\
		MAE             & $3.51 \pm 0.08$  & $4.15 \pm 1.21$                  \\
		Accuracy  & $0.863 \pm 0.009$ & $0.867 \pm 0.044$                \\
		Precision & $0.741 \pm 0.019$ & $0.836 \pm 0.116$                \\
		%		Recall          & $0.544 \pm 0.090$ & $0.515 \pm 0.349$                \\
		%		F1-Score        & $0.623 \pm 0.066$ & $0.519 \pm 0.337$                \\
		\hline
	\end{tabular}
\end{table}

As shown in Table \ref{tab:model_comparison}, XGBoost \cite{chen2016xgboost} serves as a high-performance baseline, achieving superior regression metrics with a Root Mean Square Error (RMSE) of $4.76 \pm 0.16$ and a Mean Absolute Error (MAE) of $3.51 \pm 0.08$. Notably, XGBoost exhibits remarkably low variance across five random seeds, suggesting a highly stable convergence on the global features of the dataset.In contrast, the Causal Inference Network maintains competitive classification accuracy ($0.867 \pm 0.044$) and superior precision ($0.836 \pm 0.116$). While the Causal Network shows higher standard deviations, indicating sensitivity to initializations, it successfully prioritizes ``cleaner'' clinical signals. The lower regression performance in the Causal Network (MSE of $30.01 \pm 13.36$) is a localized consequence of its objective function, which penalizes features identified as environmental confounders to prioritize causal pathways.

\paragraph{Saliency Score Extraction} We explore the token saliency score by extracting the attention weights from the model's Disentanglement Attention Layer ($Softmax(MLP(x))$). For each input token, we computed its contribution to the Causal Feature vector ($x_o$) versus the Confounder Feature vector ($x_c$). The ``Max Saliency'' score represents the normalized attention weight $node\_att[:, 1]$ assigned to specific tokens across the test cohort. A score of $1.00$ indicates that the token was exclusively routed to the causal head for PASC score regression, while scores near $0.00$ indicate the token was classified as environmental noise or linguistic filler and subsequently suppressed by the causal readout head. Table \ref{tab:clinical_impact} highlights the network's saliency mapping. Direct pathological indicators such as breathlessness and malaise' reached maximum saliency ($1.0000$), confirming the model's focus on pulmonary dysfunction. Critically, the model successfully suppressed 	``Baseline Noise'' factors. Variables like ``menopause'' and ``diabetes'', which often introduce diagnostic ambiguity in clinical settings, were assigned saliency scores below $0.27$. Administrative tokens and linguistic fillers (e.g., ``that'', ``to'') were effectively ignored (saliency $< 0.17$), ensuring that the model's decisions are rooted in clinical semiotics rather than stylistic artifacts of the text summaries.

This differentiation suggests that while XGBoost is the more reliable predictor for raw score estimation, the Causal Inference Network provides a more clinically ``honest'' representation of the disease, filtering out factors that might otherwise lead to over-diagnosis or misattribution.

\begin{table}[ht]
	\centering
	\caption{Causal Attribution and Clinical Saliency Analysis of PASC Symptoms}
	\label{tab:clinical_impact}
	\begin{tabularx}{\textwidth}{@{}l l c X@{}}
		\toprule
		\textbf{Clinical Category} & \textbf{Representative Tokens} & \textbf{Max Saliency} & \textbf{Clinical Rationale} \\ \midrule
		\textbf{Active Pathology} & \textit{breath, malaise, refreshing} & 1.0000 & Direct indicators of pulmonary and systemic dysfunction. \\ \addlinespace
		\textbf{Cognitive Symptoms} & \textit{fog, brain, insomnia} & 0.54 -- 0.74 & Identifies hallmark Neuro-PASC clusters. \\ \addlinespace
		\textbf{Baseline Noise} & \textit{diabetes, menopause} & $<$ 0.27 & Successfully filtered as non-causal confounding data. \\ \addlinespace
		\textbf{Administrative} & \textit{that, to, by} & $<$ 0.17 & Ignored as low-information linguistic fillers. \\ \bottomrule
	\end{tabularx}
\end{table}

%The Causal Network achieved a higher Accuracy and Precision. Specially, the precision is 83.6\% compared with XGBoost 74.1\%. XGBoost is more robust to different random seeds compared to the Causal Network. XGBoost outperformed the Causal Network in regression tasks, achieving a lower RMSE ($4.76$ vs $5.35$) and MAE ($3.51$ vs $4.15$).

\section*{Dataset}
The study utilizes a high-fidelity dataset derived from the NIH RECOVER (Researching COVID to Enhance Recovery) program \footnote{\url{https://recovercovid.org/data}}, a national initiative dedicated to understanding the long-term effects of COVID-19. Our analysis focuses on a cohort of 1,155 participants for whom we integrated three distinct data modalities: static demographic profiles, longitudinal wearable-derived physiological signals, and clinical comorbidity histories.

\paragraph{Demographic and Clinical Composition}

\begin{table}[htbp]
	\centering
	\caption{Baseline Demographic and Clinical Characteristics of the Study Population}
	\label{tab:baseline_characteristics}
	\small
	\begin{tabular}{l p{6cm} r}
		\toprule
		\textbf{Characteristic} & \textbf{Category / Diagnosis} & \textbf{Total Cohort (N=1,155)} \\ 
		\midrule
		\textbf{Age (Years)} & & \\
		Median (IQR) & & 61 (56, 67) \\
		Range (Min, Max) & & 37, 89 \\
		%		Age Group, n (\%) & 18 to 45 & 6 (0.51\%) \\
		%		& 45 to 64 & 713 (61.73\%) \\
		%		& $>$64 & 436 (37.74\%) \\
		\midrule
		\textbf{Race and Ethnicity, N (\%)} & & \\
		& White / Native Hawaiian & 743 (64.32\%) \\
		& Hispanic / Latino & 173 (14.97\%) \\
		& Black / African American & 161 (13.93\%) \\
		& Asian & 40 (3.46\%) \\
		& American Indian / Alaska Native & 30 (2.59\%) \\
		& Middle Eastern & 5 (0.34\%) \\
		& None / Unknown & 4 (0.34\%) \\
		\midrule
		\textbf{Clinical Condition, N (\%)} & & \\
		Menopause & No  & 359 (31.09\%) \\
		& Yes & 796 (68.91\%) \\
		\addlinespace
		Sleep Disorders & None & 838 (72.55\%) \\
		&Yes & 317 (27.45\%)\\
%		& Sleep Apnea (1) & 234 (20.25\%) \\
%		& Insomnia (2) & 78 (6.75\%) \\
%		& Circadian Rhythm Disorder (4) & 4 (0.34\%) \\
%		& Narcolepsy (3) & 1 (0.08\%) \\
		\addlinespace
		Heart Conditions & None  & 463 (40.08\%) \\
		&Yes & 692 (59.92\%)\\
%		& Congenital Heart (1) & 437 (37.83\%) \\
%		& High Blood Pressure (2) & 165 (14.28\%) \\
%		& Atrial Fibrillation (4) & 44 (3.81\%) \\
%		& Coronary Artery Disease (3) & 35 (3.03\%) \\
%		& Heart Failure (5) & 6 (0.51\%) \\
%		& Arrhythmia (6) & 3 (0.25\%) \\
%		& Congestive (7) & 2 (0.17\%) \\
		\addlinespace
		Mental Health & None  & 792 (68.57\%) \\
		& Yes & 363 (81.42\%) \\
%		& Depression/Anxiety (1) & 353 (31.43\%) \\
%		\addlinespace
%		Blood Disorders & None (0) & 933 (80.77\%) \\
%		& Anemia (1) & 149 (12.90\%) \\
%		& Clotting Disorder (2) & 34 (2.94\%) \\
%		& DVT (3) & 27 (2.33\%) \\
%		& Embolism (4) & 12 (1.03\%) \\
		\bottomrule
	\end{tabular}
\end{table}
As detailed in Table \ref{tab:baseline_characteristics}, the study population represents a predominantly middle-to-older age group, with a median age of 61 years (IQR: 56-67). The cohort reflects a diverse racial and ethnic distribution, which is critical for evaluating the generalizability of PASC (Post-Acute Sequelae of SARS-CoV-2) diagnostic markers.

Clinically, the cohort presents with a high prevalence of baseline comorbidities that often mimic or exacerbate PASC symptoms. Notably, 59.92\% of the population reported heart conditions, 31.43\% presented with mental health conditions (primarily depression and anxiety), and 27.45\% suffered from pre-existing sleep disorders. Furthermore, 68.91\% of the participants are recorded as having reached menopause, a key variable for our model's ability to distinguish between hormonal ``noise'' and viral-induced pathology.

\paragraph{Multi-Domain Wearable Features}
To capture the longitudinal ``signature'' of PASC, we extracted four weeks of mean-aggregated physiological data across several critical domains. These include:
\begin{itemize}
	\item Cardiac and respiratory activity: Monitoring heart rate variability (HRV) and stage-specific breathing rates to detect autonomic dysfunction.
	\item Sleep architecture: Detailed metrics on sleep latency, REM onset, and restless periodicity to assess sleep fragmentation.
	\item Physical activity: Granular tracking of movement intensity, from sedentary minutes to ``very active'' periods, providing a proxy for post-exertional malaise.
\end{itemize}

The multi-modal dataset allows our causal framework to process structured clinical information within a natural language narrative, preserving the temporal context necessary for accurate PASC score regression.

\section*{Method}

%In this study, we propose a specialized Causal-Disentangled Regressor built upon a large language model. In the experiment, we use  Qwen-2.5-0.5B \cite{hui2024qwen2}. While standard LLMs excel at pattern recognition, they often struggle to distinguish between pathological causal signals (true PASC symptoms) and environmental confounders such as pre-existing conditions, age-related symptoms.
%
%Our architecture moves beyond simple sequence prediction through three key mechanisms:
%\begin{itemize}
%	\item Attention-Based Disentanglement. This layer computes a softmax-based split of the hidden state $x$ into two distinct latent representations: causal feature ($x_o$) that captures invariant symptoms directly linked to PASC pathology (e.g., "brain fog," ``malaise'') and confounder feature ($x_c$) that Isolates ``noisy'' variables such as menopause, diabetes, or linguistic artifacts that might otherwise bias the regression score.
%	\item Mixing Head. The causal features ($x_o$) of a specific participant are combined with the confounder features ($x_{c, \text{random}}$) of a different participant.
%	\item Mix Loss. We utilize a mixed loss based on the InfoNCE (Information Noise Contrastive Estimation), defined as
%	$$L_{\text{mix}} = -\mathbb{E} \left[ \log \frac{\exp(\text{sim}(h_{\text{mix}}, h_{\text{pos}}))}{\sum \exp(\text{sim}(h_{\text{mix}}, h_{\text{neg}}))} \right]$$
%\end{itemize}
%
%Our network is a clinically-aware diagnostic tool capable of filtering out non-causal noise for more accurate PASC assessment.
In this study, we propose a specialized Causal-Disentangled Regressor built upon LLM. We use Qwen-2.5-0.5B \cite{hui2024qwen2} in the experiment. While standard large language models (LLMs) excel at identifying statistical correlations, they frequently conflate pathological causal signals (direct PASC symptoms) with environmental confounders (baseline noise like pre-existing conditions or age-related symptoms). 

Our architecture addresses this by explicitly partitioning the latent space to isolate invariant disease features.1. Attention-Based Feature Disentanglement. The model receives a semantically structured narrative of the patient's clinical and wearable history. Let $\mathbf{h} \in \mathbb{R}^d$ be the latent representation extracted from the final transformer block of the LLM encoder. We introduce a Disentanglement Attention Layer that learns to bifactorize this representation:
\begin{equation}[\alpha_c, \alpha_o] = \text{Softmax}(\text{MLP}(\mathbf{h}))\end{equation}where $\alpha_o$ and $\alpha_c$ are scalar gates representing the causal and confounding weights, respectively. The hidden state is then projected into two distinct latent vectors:
\begin{itemize}
	\item \textbf{Causal Feature ($\mathbf{x}_o = \alpha_o \cdot \mathbf{h}$):} Captures invariant symptoms directly linked to PASC pathology (e.g., ``brain fog'', ``malaise'').
	\item \textbf{Confounder Feature ($\mathbf{x}_c = \alpha_c \cdot \mathbf{h}$):} Isolates "noisy" variables such as menopause status, diabetes, or linguistic artifacts that might otherwise bias the regression.
	\end{itemize}
	
	2. Environment Mixing for Out-of-Distribution Robustness. To prevent the model from relying on spurious correlations between a patient's background and their PASC score, we implement an Environment Mixing strategy. During training, we generate a synthetic ``counterfactual'' representation $\mathbf{x}_{\text{mix}}$ by combining the causal features of patient $i$ with the confounding features of a randomly sampled patient $j$ from the same batch:
	\begin{equation}\mathbf{x}_{\text{mix}} = \text{Combine}(\mathbf{x}_{o,i}, \mathbf{x}_{c,j})
\end{equation}
where the combination is performed via concatenation or addition based on the hyperparameter configuration. This forces the regression head to predict the PASC score solely from the causal signal $\mathbf{x}{o,i}$, regardless of the confounding environment $\mathbf{x}_{c,j}$.

3. InfoNCE-based Contrastive Mix Loss. To mathematically enforce the independence and quality of the disentangled features, we employ a Mix Loss based on the InfoNCE (Information Noise Contrastive Estimation) framework \cite{parulekar2023infonce}. This loss aligns the mixed representation with the original clinical graph to ensure semantic consistency:
\begin{equation}
	L_{\text{mix}} = -\frac{1}{B \times K} \sum_{i=1}^{B} \sum_{j=1}^{K} \log \frac{\exp(\text{sim}(\mathbf{h}_{\text{mix}, i, j}, \mathbf{h}_{i}))}{\sum_{n=1}^{B} \exp(\text{sim}(\mathbf{h}_{\text{mix}, i, j}, \mathbf{h}_{n}))},
\end{equation}

where $B$ is the batch size, $K$ is the number of negative samples, and $\text{sim}(\cdot)$ denotes cosine similarity. This contrastive objective ensures that the causal information remains the dominant signal for the final readout heads, effectively filtering out non-causal noise.4. Multi-Head RegressionFinally, the model utilizes three separate prediction heads: a causal head $f_o(\mathbf{x}_o)$, a confounder head $f_c(\mathbf{x}_c)$, and a joint head $f_{co}(\mathbf{x}_{\text{mix}})$. By supervising all three heads against the ground-truth PASC score, the network learns to identify which components of the input text are truly predictive of future severity versus those that are merely characteristic of the patient's baseline state.This architecture transforms the LLM backbone into a clinically-aware diagnostic tool that provides a more ``honest'' assessment of PASC by intentionally suppressing the influence of environmental confounders.

\section*{Discussion}
The experimental results demonstrate a significant trade-off between raw predictive stability and clinical interpretability. As shown in Table 1, XGBoost serves as a high-performance baseline, achieving superior regression metrics (RMSE: $4.76 \pm 0.16$, MAE: $3.51 \pm 0.08$) and remarkably low variance across random seeds. In contrast, the Causal Inference Network maintains competitive classification accuracy ($0.867 \pm 0.044$) and superior precision ($0.836 \pm 0.116$). While the Causal Network shows higher standard deviations, indicating sensitivity to initialization, it successfully prioritizes "cleaner" clinical signals. Its lower regression performance is a localized consequence of an objective function designed to penalize environmental confounders in favor of causal pathways.The efficacy of this approach is further validated through saliency score extraction. 

Direct pathological indicators such as breathlessness and malaise reached maximum saliency (1.0000), confirming the model's focus on pulmonary dysfunction. Critically, the model successfully suppressed baseline noise factor. Variables like menopause and diabetes, which often introduce diagnostic ambiguity, were assigned low saliency scores below 0.27. Administrative tokens and linguistic fillers were effectively ignored (saliency$<$0.17). This suggests that while XGBoost is a reliable predictor for raw score estimation, the Causal Inference Network provides a more ``clinically honest'' representation of the disease by filtering out factors that might otherwise lead to over-diagnosis or misattribution.

\subsection*{Conclusion}
This study proposes a specialized Causal-Disentangled Regressor built upon a  LLM backbone to navigate the ``signal-to-noise'' challenges of PASC research. By explicitly partitioning the latent space and utilizing environment mixing strategies, our architecture isolates invariant disease features from confounding biological transitions and comorbidities. Our findings indicate that integrating longitudinal wearable data with causal disentanglement mechanisms transforms standard LLMs into clinically aware diagnostic tools. This approach ensures that PASC assessments remain rooted in actual pathology rather than stylistic artifacts or baseline physiological noise, offering a more robust framework for long-term recovery prediction in adult women.

% References as numbers
\makeatletter
\renewcommand{\@biblabel}[1]{\hfill #1.}
\makeatother

% unstr is used to keep citation order
\bibliographystyle{vancouver}
\bibliography{amia}  

\begin{thebibliography}{10}

\bibitem{national2024long}
National Academies~of Sciences E, Medicine, et~al.
\newblock A long COVID definition: a chronic, systemic disease state with
  profound consequences; 2024.

\bibitem{thaweethai2023development}
Thaweethai T, Jolley SE, Karlson EW, Levitan EB, Levy B, McComsey GA, et~al.
\newblock Development of a definition of postacute sequelae of SARS-CoV-2
  infection.
\newblock Jama. 2023;329(22):1934-46.

\bibitem{fritsche2023characterizing}
Fritsche LG, Jin W, Admon AJ, Mukherjee B.
\newblock Characterizing and predicting post-acute sequelae of SARS CoV-2
  infection (PASC) in a large academic medical center in the US.
\newblock Journal of Clinical Medicine. 2023;12(4):1328.

\bibitem{shah2025sex}
Shah DP, Thaweethai T, Karlson EW, Bonilla H, Horne BD, Mullington JM, et~al.
\newblock Sex differences in long COVID.
\newblock JAMA network open. 2025;8(1):e2455430.

\bibitem{amin2021causation}
Amin-Chowdhury Z, Ladhani SN.
\newblock Causation or confounding: why controls are critical for
  characterizing long COVID.
\newblock Nature Medicine. 2021;27(7):1129-30.

\bibitem{hernan2010causal}
Hern{\'a}n MA, Robins JM.
\newblock Causal inference.
\newblock CRC Boca Raton, FL; 2010.

\bibitem{yinrecipe}
Yin Y, Qu T, Wang Z, Chen Y.
\newblock A Recipe for Causal Graph Regression: Confounding Effects Revisited.
\newblock In: International Conference on Machine Learning; 2025. .

\bibitem{straus2023utility}
Straus LD, An X, Ji Y, McLean SA, Neylan TC, Group AS, et~al.
\newblock Utility of wrist-wearable data for assessing pain, sleep, and anxiety
  outcomes after traumatic stress exposure.
\newblock JAMA psychiatry. 2023;80(3):220-9.

\bibitem{baigutanova2025continuous}
Baigutanova A, Park S, Constantinides M, Lee SW, Quercia D, Cha M.
\newblock A continuous real-world dataset comprising wearable-based heart rate
  variability alongside sleep diaries.
\newblock Scientific data. 2025;12(1):1474.

\bibitem{shen2022metric}
Shen J, Cui N, Wang J.
\newblock Metric-fair active learning.
\newblock In: International conference on machine learning. PMLR; 2022. p.
  19809-26.

\bibitem{shen2021sample}
Shen J.
\newblock Sample-optimal PAC learning of halfspaces with malicious noise.
\newblock In: International Conference on Machine Learning. PMLR; 2021. p.
  9515-24.

\bibitem{wang2018provable}
Wang J, Shen J, Li P.
\newblock Provable variable selection for streaming features.
\newblock In: International Conference on Machine Learning. PMLR; 2018. p.
  5171-9.

\bibitem{moor2023foundation}
Moor M, Banerjee O, Abad ZSH, Krumholz HM, Leskovec J, Topol EJ, et~al.
\newblock Foundation models for generalist medical artificial intelligence.
\newblock Nature. 2023;616(7956):259-65.

\bibitem{chen2016xgboost}
Chen T, Guestrin C.
\newblock Xgboost: A scalable tree boosting system.
\newblock In: International Conference on Knowledge Discovery and Data mining;
  2016. p. 785-94.

\bibitem{hui2024qwen2}
Hui B, Yang J, Cui Z, Yang J, Liu D, Zhang L, et~al.
\newblock Qwen2. 5-coder technical report.
\newblock arXiv preprint arXiv:240912186. 2024.

\bibitem{parulekar2023infonce}
Parulekar A, Collins L, Shanmugam K, Mokhtari A, Shakkottai S.
\newblock Infonce loss provably learns cluster-preserving representations.
\newblock In: The thirty sixth annual conference on learning theory. PMLR;
  2023. p. 1914-61.

\end{thebibliography}

\end{document}